\documentclass[letterpaper,10pt,conference]{ieeeconf}
\IEEEoverridecommandlockouts
\usepackage{cuted}
\usepackage{comment}
\usepackage{cite}
\usepackage{amsmath,amssymb,amsfonts}
\usepackage{algorithmic}
\usepackage{graphicx}
\usepackage{float}
\usepackage{booktabs}
\usepackage{textcomp}
\usepackage{xcolor}
\usepackage{hyperref}
\def\BibTeX{{\rm B\kern-.05em{\sc i\kern-.025em b}\kern-.08em
    T\kern-.1667em\lower.7ex\hbox{E}\kern-.125emX}}
\begin{document}
\title{ForeTac-VLA: A Forecasting-Based Tactile-Vision-Language-Action Model
for Contact-Rich Robotic Manipulation}

\author{
Zhengyu Tao,
Xin Li,
and Xin Wang$^{*}$%
\thanks{Zhengyu Tao, Xin Li, and Xin Wang are with Texas A\&M University,
College Station, TX 77843, USA
(e-mails: zhengyutao@tamu.edu; xinli@tamu.edu; xin.wang@tamu.edu).}
\thanks{$^{*}$Corresponding author: Xin Wang.}
}

\maketitle

\vspace{-1.8cm}

\begin{strip}
    \vspace{-1.0cm}
    \centering

    \includegraphics[width=0.88\textwidth]{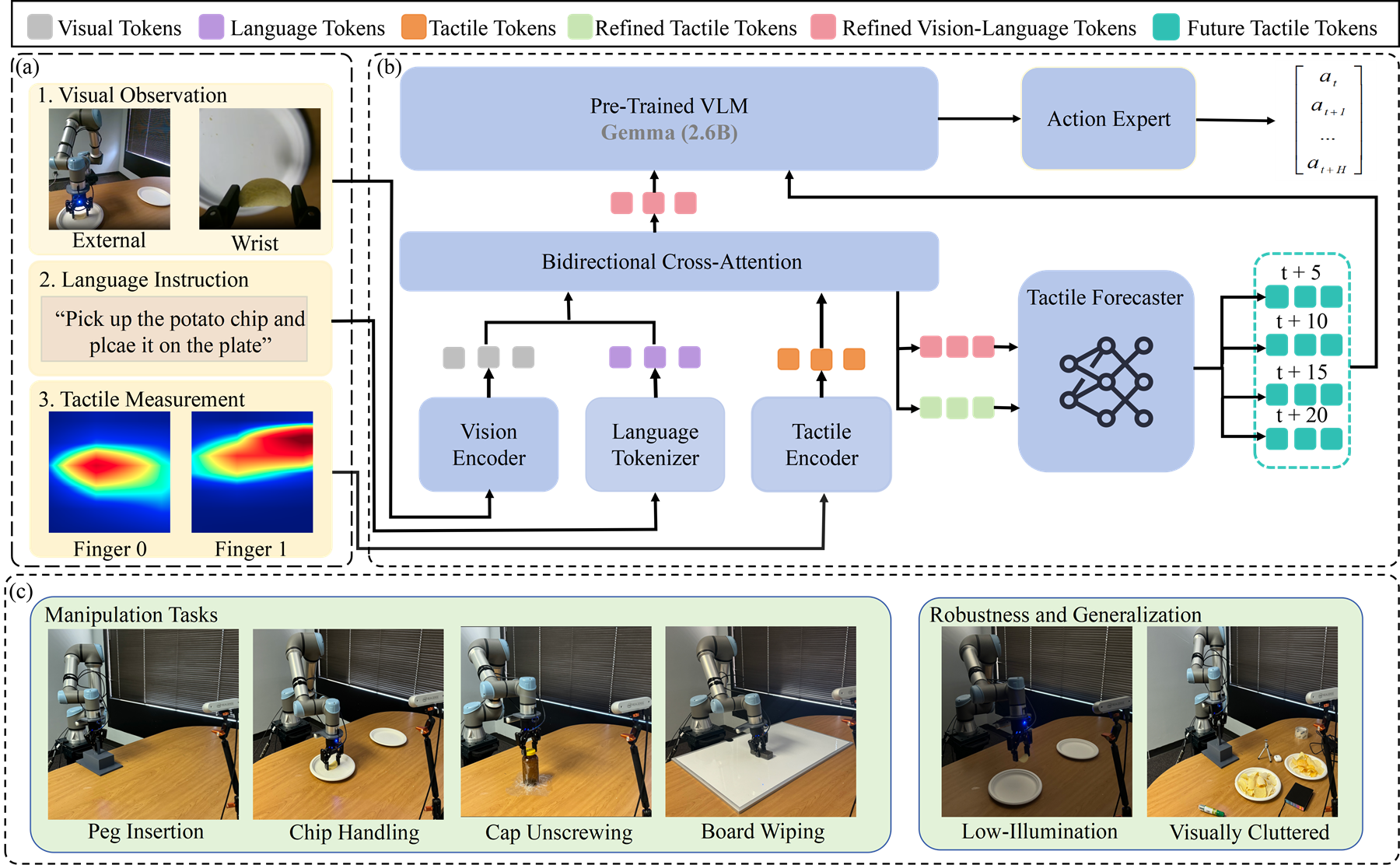}

    \vspace{1mm}

    \refstepcounter{figure}

    \begin{minipage}{0.98\textwidth}
    \small
    Fig.~\thefigure. \textbf{Overview of ForeTac-VLA.}
    (a) Multimodal observations including external and wrist-view RGB images, a language instruction, and tactile measurements from two fingertips.
    (b) ForeTac-VLA architecture. Modality-specific encoders produce visual-language tokens and tactile tokens, which interact through bidirectional cross-attention. A tactile forecaster predicts multi-step future tactile states, whose tokens are combined with refined vision-language tokens to condition action generation.
    (c) Four contact-rich manipulation tasks and robustness evaluations under low-illumination and visually cluttered environments.
    \end{minipage}

    \label{fig:architecture}
    \vspace{0.2cm}
\end{strip}

\begin{abstract}
Vision-language-action (VLA) models have demonstrated strong capabilities in robotic manipulation, yet their reliance on visual perception limits robustness in contact-rich environments, where critical physical interaction states may not be visually observable. 
Existing tactile-enhanced VLA methods improve physical grounding using observed tactile feedback, but most remain largely reactive rather than explicitly modeling how contact may evolve. 
Therefore, we propose \textbf{ForeTac-VLA}, a forecasting-based tactile-vision-language fusion model that predicts future tactile states to guide action generation. Specifically, ForeTac-VLA encodes recent tactile observations into temporal representations and integrates them with vision-language features through bidirectional cross-attention. Further, a transformer-based forecasting module predicts multi-step future tactile states, enabling the model to reason jointly over observed and anticipated contact. Finally, the fused multimodal representations and predicted future tactile states are fed into the VLA backbone to condition action generation. To stabilize training, a ground-truth-to-prediction curriculum is employed when early forecasts are unreliable. Across four real-world contact-rich manipulation tasks, ForeTac-VLA achieves an average success rate of 95\%, outperforming the fine-tuned VLA model by 36.25 percentage points and state-of-the-art tactile-enhanced VLA baselines by over 22 percentage points. 
ForeTac-VLA also maintains strong performance under low-illumination and visually cluttered conditions.
Video demonstrations can be found on \href{https://foretac-vla.github.io/}{https://foretac-vla.github.io/}.
\end{abstract}

\section{INTRODUCTION}
Recent years have witnessed rapid advances in vision-language-action (VLA) models for robotic manipulation, which leverage large-scale pretrained vision-language representations to support semantic understanding, task generalization, and action generation~\cite{rt2,openvla,pi0,pi05}. However, most existing VLA models remain predominantly vision-driven and often struggle in contact-rich settings. In tasks such as constrained
insertion, fragile-object handling, sustained surface interaction, and
rotational manipulation, visually similar observations can correspond to substantially different physical states. An object may appear properly aligned while being mechanically jammed, a fragile object may appear securely grasped while experiencing excessive contact pressure, or a grasp may appear stable while contact is gradually weakening. These physical differences are difficult to infer from images alone but directly determine whether the robot should continue, adjust, release, or terminate an action. Therefore, tactile feedback provides critical complementary information for improving the precision and robustness of VLA models in contact-rich manipulation~\cite{vtla,tacvla,taccorl,torlvla}.

Prior work has explored several strategies for incorporating tactile sensing into VLA models which can be termed as TacVLA models. They include token concatenation~\cite{tacvla,tactilevla,vtla}, feature-wise modulation~\cite{tacfilm}, cross-attention-based fusion~\cite{atvla}, and semantically aligned tactile representation learning~\cite{omnivtla}. While these approaches demonstrate the benefits of tactile feedback for contact-rich manipulation, they primarily focus on integrating observed tactile information, with limited attention to bidirectional tactile-vision interaction and future contact evolution. Consequently, existing models remain largely reactive to physical interaction rather than explicitly anticipating how contact will evolve.

We address this gap by proposing \textbf{ForeTac-VLA}, a forecasting-based tactile-VLA model that forecasts future tactile states for contact-rich robotic manipulation. At each control step, ForeTac-VLA receives visual observations, a language instruction, robot proprioception, and a short sequence of tactile measurements. The tactile sequence is encoded into temporal tactile tokens, which interact with vision-language representations through bidirectional cross-attention. Based on the fused representation, a transformer-based forecasting module then predicts multi-step future tactile states. Finally, the fused multimodal representation and predicted future tactile states jointly condition action generation, enabling the model to anticipate upcoming contact changes rather than relying solely on observed tactile feedback. To prevent unreliable early predictions from destabilizing model learning, we employ a two-stage ground-truth-to-prediction conditioning curriculum, which first conditions action generation on ground-truth future tactile states and then switches to model-predicted states.

To evaluate ForeTac-VLA, we conduct real-robot experiments on four
contact-rich manipulation tasks: peg insertion, chip handling, cap unscrewing, and board wiping. ForeTac-VLA achieves an average success rate of 95\%, compared with 58.75\% for the fine-tuned VLA model, 71.25\% for the FiLM-fusion TacVLA model and 72.5\% for the concat-and-gating TacVLA model. It achieves the highest success rate across all four tasks, demonstrating consistent improvements across diverse contact dynamics. Ablation studies evaluate the contribution of tactile input, assess the effect of future tactile predictions, and compare different tactile-vision-language fusion strategies on model performance. We further evaluate robustness and generalization under challenging conditions, including low illumination and visually cluttered environments.

The main contributions of this work are summarized as follows:
\begin{itemize}
    \item We introduce ForeTac-VLA, a forecasting-based tactile VLA model that explicitly predicts future tactile states and uses them to guide action generation in contact-rich manipulation.

    \item We design a multimodal strategy that enables bidirectional interaction between tactile and vision-language representations and leverages a transformer-based forecasting module to predict multi-step future tactile states.

    \item We evaluate ForeTac-VLA on four real-world contact-rich tasks with an average success rate of 95\%  and demonstrate robustness and generalization under challenging conditions.
\end{itemize}

\begin{figure}[H]
    \centering
    \includegraphics[width=0.75\columnwidth]{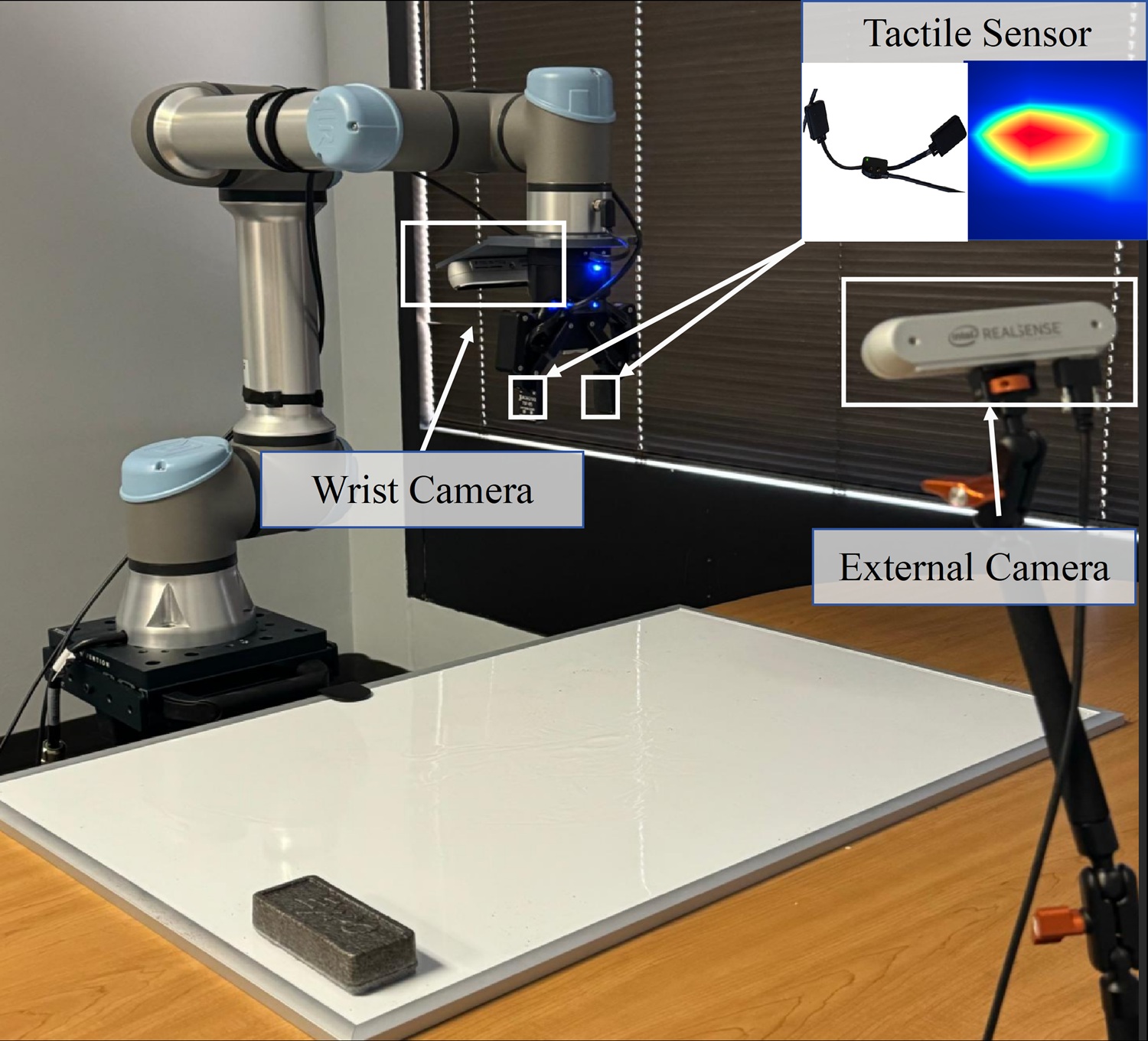}
    \caption{Real-robot experimental setup. The UR7e manipulator is equipped with a tactile-enabled parallel gripper, a wrist-mounted camera, and a fixed external camera.}
    \label{fig:setup}
\end{figure}

\section{RELATED WORK}

\begin{figure*}[!t]
    \centering
    \includegraphics[width=0.8\textwidth]{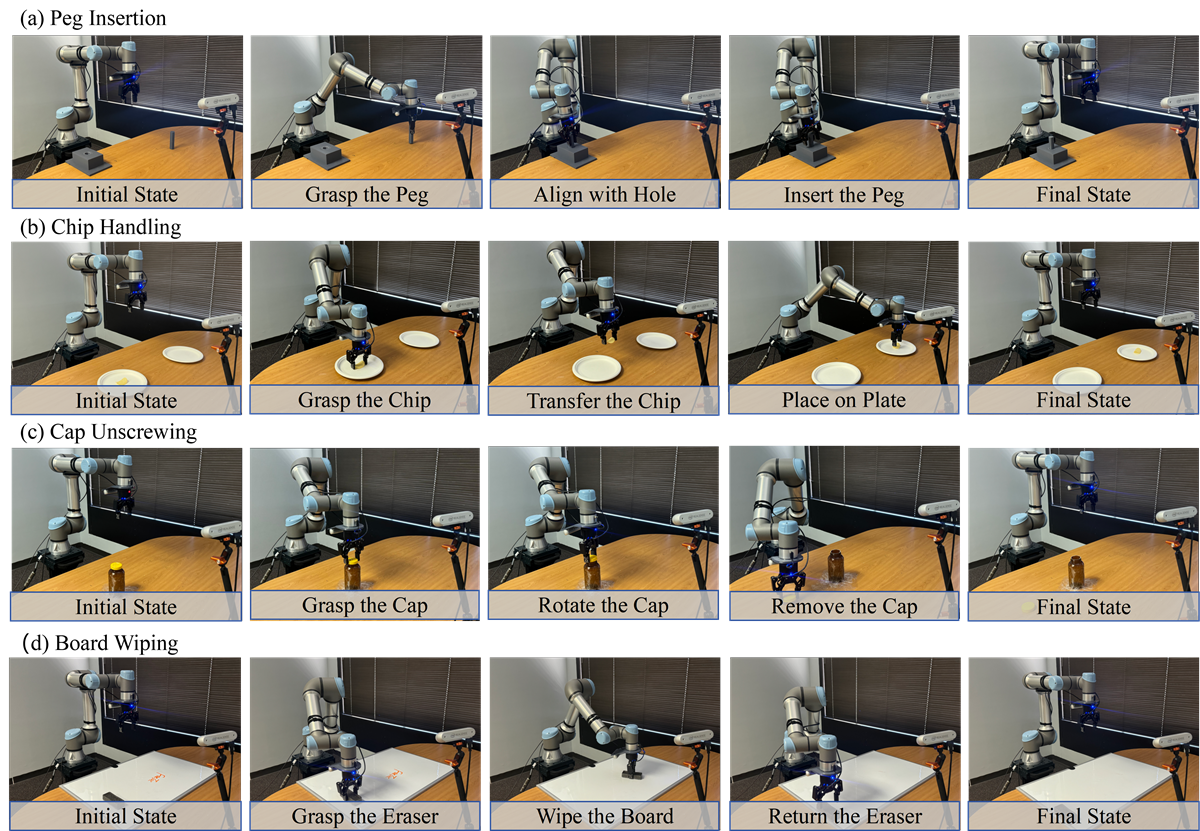}
    \caption{Representative execution sequences of four contact-rich manipulation tasks: (a) peg insertion, (b) chip handling, (c) cap unscrewing, and (d) board wiping. Each row illustrates the task progression from the initial state through key contact-rich interactions to successful completion.}
    \label{fig:tasks}
\end{figure*}

\subsection{Tactile Sensing for Robotic Manipulation}

\begin{table*}[!t]
\centering
\caption{Success rate comparison on real-world contact-rich manipulation tasks. Each model is evaluated over 20 trials per task.}
\label{tab:main_results}

\small
\setlength{\tabcolsep}{8pt}
\renewcommand{\arraystretch}{1.15}

\begin{tabular}{lccccc}
\toprule
Method & Peg Insertion & Chip Handling & Cap Unscrewing & Board Wiping & Average \\
\midrule
ACT & 2/20 & 5/20 & 0/20 & 4/20 & 13.75\% \\
Fine-tuned VLA & 13/20 & 11/20 & 12/20 & 11/20 & 58.75\% \\
FiLM-Fusion TacVLA & 14/20 & 14/20 & 16/20 & 13/20 & 71.25\% \\
Concat-and-Gating TacVLA & 16/20 & 13/20 & 15/20 & 14/20 & 72.50\% \\
\textbf{ForeTac-VLA (Ours)} & \textbf{20/20} & \textbf{19/20} & \textbf{19/20} & \textbf{18/20} & \textbf{95.00\%} \\
\bottomrule
\end{tabular}
\end{table*}

Tactile sensing has been explored for contact-rich robotic manipulation because visual observations alone often provide insufficient information about physical interaction~\cite{evetac,tactilealoha,fingersts,simultaneous_tactile_visual,hu2025dexterous}. Different tactile sensing technologies provide distinct forms of information about physical contact. Vision-based tactile sensors, such as GelSight and DIGIT~\cite{gelsight,digitac}, capture fine-grained contact geometry, deformation, and slip, whereas force- and pressure-based tactile arrays~\cite{capacitive_tactile} measure contact forces or pressure distributions. These tactile sensing modalities provide physical information that complements visual perception in contact-rich manipulation.

Prior work has incorporated tactile feedback into a broad range of robotic manipulation tasks, including grasping, insertion, surface interaction, and dexterous manipulation~\cite{vitamin,vital,3dvitac,polytouch}. More recently, learning-based methods have developed tactile-vision manipulation policies using imitation learning, reinforcement learning, and diffusion models to address contact-rich manipulation tasks~\cite{3dvitac,reactive,bitouch,tacdiffusion}. These methods demonstrate the benefits of tactile sensing for fragile-object handling, dexterous manipulation, high-precision insertion, and closed-loop contact control. Beyond reactive tactile feedback,~\cite{vitacformer} has also explored predictive tactile-vision representation learning to model future contact dynamics. However, these methods are primarily designed for task-specific manipulation, limiting their applicability to general-purpose robotic manipulation.

Our work employs compact pressure-sensitive tactile arrays to capture physical interaction states and integrates these measurements into a general-purpose robotic manipulation framework for contact-rich tasks.

\subsection{Vision-Language-Action Models}

VLA models provide a general framework for connecting visual perception, language understanding, and robotic control. Representative models such as RT-2~\cite{rt2}, OpenVLA~\cite{openvla}, and $\pi_{0.5}$~\cite{pi05} leverage large-scale pretrained vision-language representations and robot demonstrations to support semantic generalization and action generation across diverse manipulation tasks. Despite these advances, most existing VLA models remain predominantly vision-driven and lack direct access to the fine-grained physical interaction states required for contact-rich manipulation. This limitation has motivated the integration of tactile sensing into VLA models.

Our work extends the VLA models with predictive temporal tactile feedback, allowing action generation to account for physical contact states that are difficult to infer from vision and language alone.

\subsection{Tactile-Enhanced Vision-Language-Action Models}

Very recent studies have incorporated tactile sensing into VLA models through various multimodal integration strategies, including token-based fusion~\cite{tactilevla,tacvla,vtla}, feature-wise modulation~\cite{tacfilm}, cross-attention-based fusion~\cite{atvla}, and tactile representation learning~\cite{omnivtla}. These studies demonstrate that tactile feedback can improve physical grounding and robustness in contact-rich manipulation. Further, predictive modeling of tactile interaction is gaining increasing attention as a means of capturing the temporal evolution of physical contact. For example, UniTacVLA~\cite{unitacvla} leverages current tactile information and future tactile predictions to refine generated actions. However, how to effectively integrate historical and future temporal tactile dynamics with vision-language representations to directly guide action generation remains challenging.

To this end, we propose ForeTac-VLA, which explicitly encodes tactile history, enables multi-step future tactile forecasts, and directly conditions the action generation process.

\section{METHOD}
In this section, we first introduce the overall architecture of ForeTac-VLA (Sec.~III-A), followed by the bidirectional cross-attention module for multimodal fusion (Sec.~III-B). We then present the transformer-based multi-step future tactile forecasting module (Sec.~III-C). Finally, we describe the training procedure in Sec.~III-D.

\begin{figure}[H]
    \centering
    \includegraphics[width=\columnwidth]{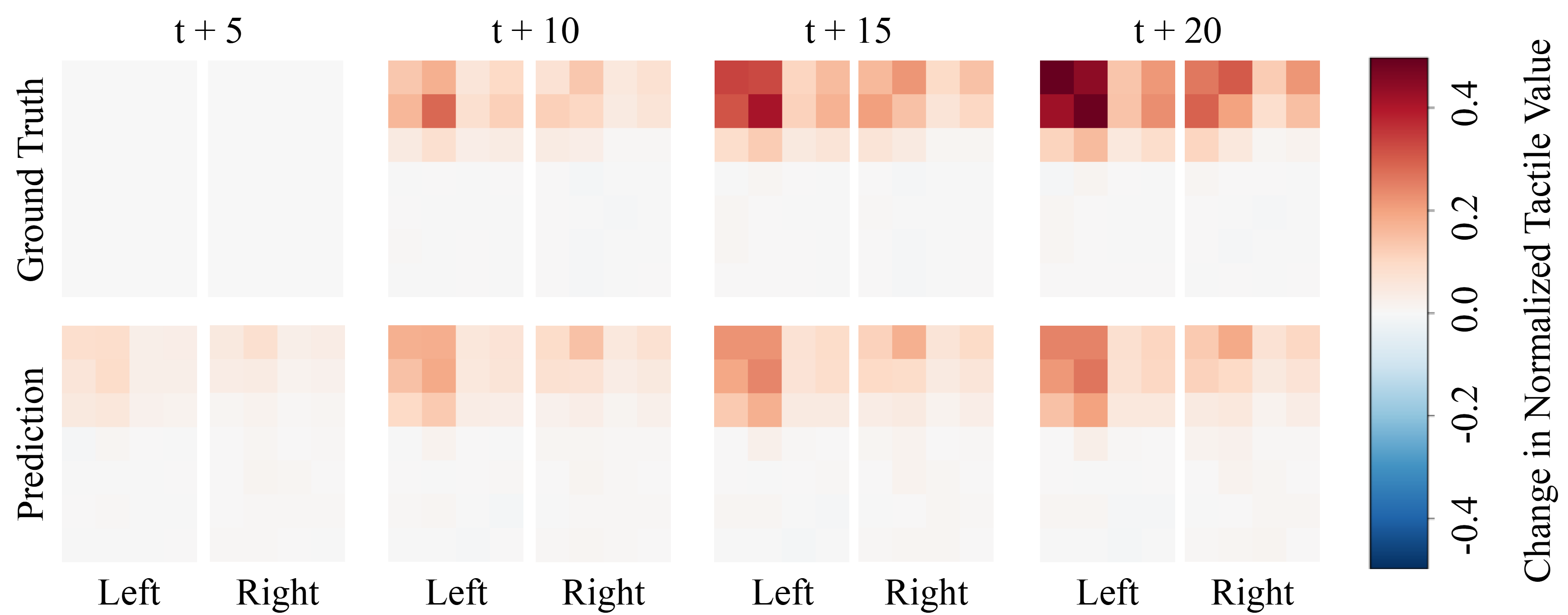}
    \caption{Qualitative comparison of predicted and ground-truth future tactile changes during a representative contact transition. Results are shown for the left and right tactile sensors at four future horizons, $t+5$, $t+10$, $t+15$, and $t+20$, corresponding to approximately 83, 167, 250, and 333 ms, respectively.}
    \label{fig:tactile_forecasting}
\end{figure}

\subsection{Overall Model Architecture}

\begin{figure*}[t]
    \centering
    \includegraphics[width=0.85\textwidth]{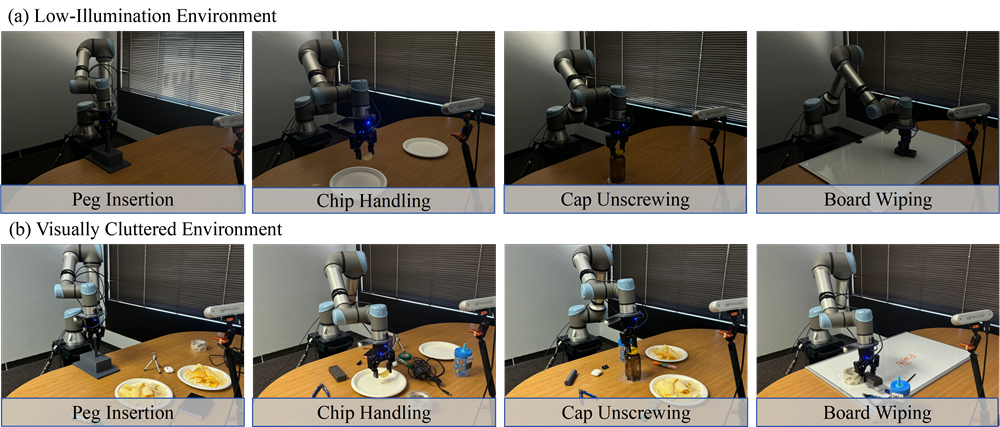}
    \caption{Robustness evaluation environments for four contact-rich manipulation tasks. The top row shows the low-illumination environment, and the bottom row shows the visually cluttered environment with task-irrelevant objects introduced into the workspace.}
    \label{fig:robustness_env}
\end{figure*}

As illustrated in Fig.~\ref{fig:architecture}, ForeTac-VLA is built upon the pretrained $\pi_{0.5}$ VLA model and extends it with temporal tactile encoding, bidirectional cross-attention, and multi-step future tactile forecasting. At each control step $t$, the model receives visual observations from the wrist and external cameras, a language instruction, robot proprioception, and tactile measurements from the most recent $K = 10$ time steps.

Following the $\pi_{0.5}$ architecture, visual observations are encoded by the pretrained SigLIP vision encoder~\cite{siglip}, while the language instruction and discretized robot proprioception are embedded through the PaliGemma tokenizer~\cite{beyer2024paligemma}. The resulting vision-language token sequence is defined as
\begin{equation}
    z_t^{\mathrm{VL}}
    =
    \left[
    z_t^{\mathrm{vis}};
    z_t^{\mathrm{lan+pro}}
    \right],
\end{equation}
where $z_t^{\mathrm{vis}}$ and $z_t^{\mathrm{lan+pro}}$ denote the visual tokens and the language plus robot proprioception tokens, respectively.

For tactile input, each gripper finger provides a $7\times4$ tactile array at each time step. We use the tactile measurements from the most recent $K$ time steps to capture short-term contact dynamics. The absolute tactile history and its changes relative to the first frame are concatenated and flattened, and then processed by a lightweight Multi-Layer Perceptron (MLP)-based tactile encoder to obtain the tactile tokens $z_t^{\mathrm{Tac}}$.

Given $z_t^{\mathrm{Tac}}$ and $z_t^{\mathrm{VL}}$, bidirectional cross-attention produces refined tactile and vision-language tokens, denoted by $z_t^{\mathrm{Tac,r}}$ and $z_t^{\mathrm{VL,r}}$, respectively. Both refined token sequences are fed into a transformer-based tactile forecaster to predict tactile states at multiple future time steps. The predicted future tactile states are subsequently encoded into future tactile tokens $z_t^{\mathrm{future}}$. The refined vision-language tokens and future tactile tokens are then concatenated to form the multimodal prefix
\begin{equation}
    z_t^{\mathrm{prefix}}
    =
    \left[
    z_t^{\mathrm{VL,r}};
    z_t^{\mathrm{future}}
    \right].
\end{equation}

Following the original $\pi_{0.5}$ architecture, the multimodal prefix $z_t^{\mathrm{prefix}}$ conditions the action expert, which generates continuous action chunks using a flow-matching objective. In this way, bidirectional cross-attention provides refined multimodal representations for both future tactile forecasting and action generation, while the predicted future tactile states further condition action generation on upcoming contact dynamics.

\begin{table*}[t]
\centering
\caption{Ablation study of ForeTac-VLA. Each model variant is evaluated over 20 trials per task. The best results are shown in bold. }
\label{tab:ablation}
\renewcommand{\arraystretch}{1.15}
\setlength{\tabcolsep}{10pt}
\begin{tabular}{lccccc}
\hline
\textbf{Model Variant}
& \textbf{Peg Insertion}
& \textbf{Chip Handling}
& \textbf{Cap Unscrewing}
& \textbf{Board Wiping}
& \textbf{Average} \\
\hline

w/o Tactile
& 11/20
& 13/20
& 12/20
& 11/20
& 58.75\% \\

w/ Tactile Concatenation
& 12/20
& 14/20
& 14/20
& 15/20
& 68.75\% \\

w/ Single/Bidirectional-Cross
& 15/20
& 17/20
& 16/20
& 16/20
& 80.00\% \\

w/ Single-Cross + Forecasting
& 14/20
& 18/20
& 17/20
& 16/20
& 81.25\% \\

\textbf{ForeTac-VLA (Ours)}
& \textbf{20/20}
& \textbf{19/20}
& \textbf{19/20}
& \textbf{18/20}
& \textbf{95.00\%} \\
\hline
\end{tabular}
\end{table*}

\subsection{Bidirectional Cross-Attention-Based Multimodal Fusion}

The vision-language tokens $z_t^{\mathrm{VL}}$ encode visual information, language instruction, and robot proprioception, while the tactile tokens $z_t^{\mathrm{Tac}}$ capture physical information about contact interactions. Simply concatenating the two token sequences does not explicitly model their cross-modal dependencies. We therefore introduce bidirectional cross-attention to enable information exchange between tactile and vision-language representations.

To incorporate vision-language context into the tactile tokens, $z_t^{\mathrm{Tac}}$ attends to $z_t^{\mathrm{VL}}$:
\begin{equation}
    z_t^{\mathrm{Tac,r}}
    =
    \operatorname{CrossAttn}
    \left(
    Q=z_t^{\mathrm{Tac}},
    K=z_t^{\mathrm{VL}},
    V=z_t^{\mathrm{VL}}
    \right).
\end{equation}
Conversely, to incorporate physical contact information into the vision-language tokens, $z_t^{\mathrm{VL}}$ attends to $z_t^{\mathrm{Tac}}$:
\begin{equation}
    z_t^{\mathrm{VL,r}}
    =
    \operatorname{CrossAttn}
    \left(
    Q=z_t^{\mathrm{VL}},
    K=z_t^{\mathrm{Tac}},
    V=z_t^{\mathrm{Tac}}
    \right),
\end{equation}
where $\operatorname{CrossAttn}(\cdot)$ denotes the multi-head cross-attention-based refinement operation.

This bidirectional interaction allows the tactile tokens to incorporate vision-language context while simultaneously refining the vision-language tokens with physical contact information. The resulting $z_t^{\mathrm{Tac,r}}$ and $z_t^{\mathrm{VL,r}}$ provide refined multimodal representations for future tactile forecasting, while $z_t^{\mathrm{VL,r}}$ also contributes directly to the multimodal prefix used for action generation.

\subsection{Transformer-Based Multi-Step Tactile Forecasting}

Current tactile observations describe physical interactions that have already occurred, whereas action generation can benefit from information about how contact will evolve over the upcoming control horizon. We therefore introduce a transformer-based tactile forecasting module that predicts multi-step future tactile states from the refined tactile and vision-language tokens.

Given $z_t^{\mathrm{Tac,r}}$ and $z_t^{\mathrm{VL,r}}$, we concatenate the two refined token sequences to form the multimodal memory of the tactile forecaster,
\begin{equation}
    z_t^{\mathrm{mem}}
    =
    \left[
    z_t^{\mathrm{Tac,r}};
    z_t^{\mathrm{VL,r}}
    \right].
\end{equation}
The tactile forecaster predicts tactile-state changes at multiple future temporal offsets,
\begin{equation}
    \widehat{\Delta T}_{t+\mathcal{O}}
    =
    F_{\mathrm{pred}}
    \left(
    z_t^{\mathrm{mem}}
    \right),
\end{equation}
where $T_t$ denotes the tactile state at dataset time step $t$ and
$\mathcal{O}=\{5,10,15,20\}$ denotes the set of future prediction offsets. The training trajectories are sampled at 60 Hz, corresponding to nominal prediction horizons of approximately 0.083, 0.167, 0.250, and 0.333 s, respectively. The corresponding future tactile states are reconstructed as
\begin{equation}
    \widehat{T}_{t+\mathcal{O}}
    =
    T_t+\widehat{\Delta T}_{t+\mathcal{O}}.
\end{equation}

The predicted future tactile states are subsequently encoded into future tactile tokens $z_t^{\mathrm{future}}$, which are combined with the refined vision-language tokens to condition action generation. In this way, ForeTac-VLA directly incorporates predicted future tactile states into action generation rather than using future tactile forecasting solely as an auxiliary prediction objective.
\subsection{Training Procedure}
ForeTac-VLA is initialized from the pretrained $\pi_{0.5}$ checkpoint and fine-tuned using Low-Rank Adaptation (LoRA). During training, the newly introduced tactile encoder, bidirectional cross-attention module, and tactile forecaster are optimized jointly with the trainable components of the VLA backbone, while the remaining pretrained parameters are kept frozen.

To stabilize tactile-conditioned action generation, we adopt a
two-stage conditioning curriculum. During the first 75\% of training steps, ground-truth future tactile tokens are used for action conditioning, allowing the policy to learn from accurate future tactile measurements. During the remaining 25\%, they are replaced by the predicted future tactile tokens, exposing the policy to the forecasting errors encountered during inference. This curriculum reduces the discrepancy between training and inference and enables action generation to better utilize predicted future tactile dynamics.

\section{EXPERIMENTS}
In this section, the experiments are presented across five aspects: (1) \textbf{Experimental Setup}: the real-robot platform, manipulation tasks, data collection and training procedures, and evaluation approach and metrics; (2) \textbf{Experimental Results}: comparisons with the fine-tuned VLA model and state-of-the-art tactile-enhanced VLA models; (3) \textbf{Ablation Study}: the contributions of tactile input, future tactile forecasting, and multimodal fusion; (4) \textbf{Robustness and Generalization}: performance under challenging visual conditions, including low-illumination and visually cluttered environments; and (5) \textbf{Failure Case Analysis}: representative failure modes observed in all the VLA models across the manipulation tasks.

\subsection{Experimental Setup}
\textbf{Experimental Platform.}
Our real-robot experimental platform is shown in Fig.~\ref{fig:setup}. It consists of a UR7e robotic arm equipped with a Robotiq 2F-85 parallel gripper and TSF-85 tactile sensors mounted on both fingers. Visual observations are captured by two Intel RealSense D455f cameras: a fixed external camera provides a global view of the workspace, while a wrist-mounted camera provides close-range observations during manipulation. The tactile sensors provide tactile measurements from both fingers, complementing visual observations during physical interaction.

\textbf{Manipulation Tasks.}
We evaluate ForeTac-VLA on four real-world contact-rich manipulation tasks, as shown in Fig.~\ref{fig:tasks}: \textit{Peg Insertion}, \textit{Chip Handling}, \textit{Cap Unscrewing}, and \textit{Board Wiping}. Each task involves a distinct form of physical interaction, including precise contact alignment, stable grasping of fragile objects, rotational contact, and sustained surface contact. Fig.~\ref{fig:tasks} illustrates representative execution sequences from the initial state through contact-rich interaction to successful task completion.

\textbf{Data Collection and Training.}
We collect 280 teleoperated demonstrations across the four tasks, with 70 demonstrations per task, using a GELLO-based teleoperation interface~\cite{Gello}. During data collection, visual observations, tactile measurements, robot proprioception, and actions are synchronously recorded at 60 Hz. ForeTac-VLA is initialized from the pretrained $\pi_{0.5}$ checkpoint and fine-tuned on the collected demonstrations together with the proposed bidirectional cross-attention and future tactile forecasting modules. For a fair comparison, all models are trained for 60,000 optimization steps using the same dataset. The training resources include an Intel Core Ultra 9 285K Central Processing Unit (CPU), 128 GB memory, and an NVIDIA RTX PRO 6000 Blackwell @ 96 GB Graphic Processing Unit (GPU).

\textbf{Evaluation Approach and Metrics.}
We compare ForeTac-VLA with four baseline models: (1) a fine-tuned VLA model without tactile input; (2) a FiLM-fusion TacVLA model based on the fusion strategy in prior work~\cite{tacfilm}, where tactile representations condition the visual representations through feature-wise linear modulation; (3) a concat-and-gating TacVLA model based on the fusion strategy in prior work~\cite{tacvla}, where tactile representations are concatenated with the VLA representations and modulated by a contact-aware gating mechanism; and (4) ACT~\cite{zhao2023learning}, a transformer-based imitation learning model. All models are evaluated under the same experimental conditions. We use task success rate as the evaluation metric. Each model is evaluated over 20 independent trials for each of the four tasks, resulting in 80 trials per model.

\subsection{Experimental Results}

Table~\ref{tab:main_results} reports the quantitative performance of ForeTac-VLA and the baseline models across the four contact-rich manipulation tasks. ForeTac-VLA achieves the best overall performance, with an average success rate of 95.0\%, compared with 13.75\% for ACT, 58.75\% for the fine-tuned VLA model, 71.25\% for the FiLM-fusion TacVLA model, and 72.50\% for the concat-and-gating TacVLA model. This corresponds to an improvement of 36.25 percentage points over the fine-tuned VLA model, 23.75 percentage points over the FiLM-fusion TacVLA model, and 22.50 percentage points over the concat-and-gating TacVLA model. Both tactile-enhanced baseline models substantially outperform the fine-tuned VLA model, confirming the benefit of tactile feedback for contact-rich manipulation. However, the remaining performance gap to ForeTac-VLA suggests that existing tactile integration strategies are still limited in capturing multimodal interactions and anticipating future contact evolution.

ForeTac-VLA also achieves the highest success rate on every task, reaching 20/20 (100\%) on \textit{Peg Insertion}, 19/20 (95\%) on \textit{Chip Handling}, 19/20 (95\%) on \textit{Cap Unscrewing}, and 18/20 (90\%) on \textit{Board Wiping}. The consistent gains across all four tasks indicate that bidirectional cross-attention enables more effective interaction between tactile and vision-language representations, while future tactile forecasting provides predicted future tactile states for action generation. The performance gap is particularly evident in tasks requiring precise contact regulation and sustained physical interaction, where accurate estimation of the underlying contact state is critical. Together, these results demonstrate that integrating tactile information with bidirectional cross-attention and future tactile forecasting improves the reliability of VLA models across diverse contact-rich manipulation tasks.

\subsection{Ablation Study}
To investigate the contribution of each component in ForeTac-VLA, we conduct a systematic ablation study on tactile input, future tactile forecasting, and multimodal fusion. As shown in Table~\ref{tab:ablation}, we progressively introduce these components while keeping the remaining training and evaluation settings unchanged. Each model variant is evaluated over 20 trials on each of the four contact-rich tasks.

\textbf{1) Effect of Tactile Input:}
Introducing tactile measurements through direct token concatenation improves the average success rate from 58.75\% to 68.75\%, demonstrating the benefit of tactile feedback for contact-rich manipulation. Replacing direct concatenation with single-direction cross-attention further increases the average success rate to 80.00\%, indicating more effective interaction between tactile and vision-language representations. Without future tactile forecasting, bidirectional cross-attention is equivalent to single-direction cross-attention in our implementation and is therefore not reported separately.

\begin{figure}[!t]
    \centering
    \includegraphics[width=\columnwidth]{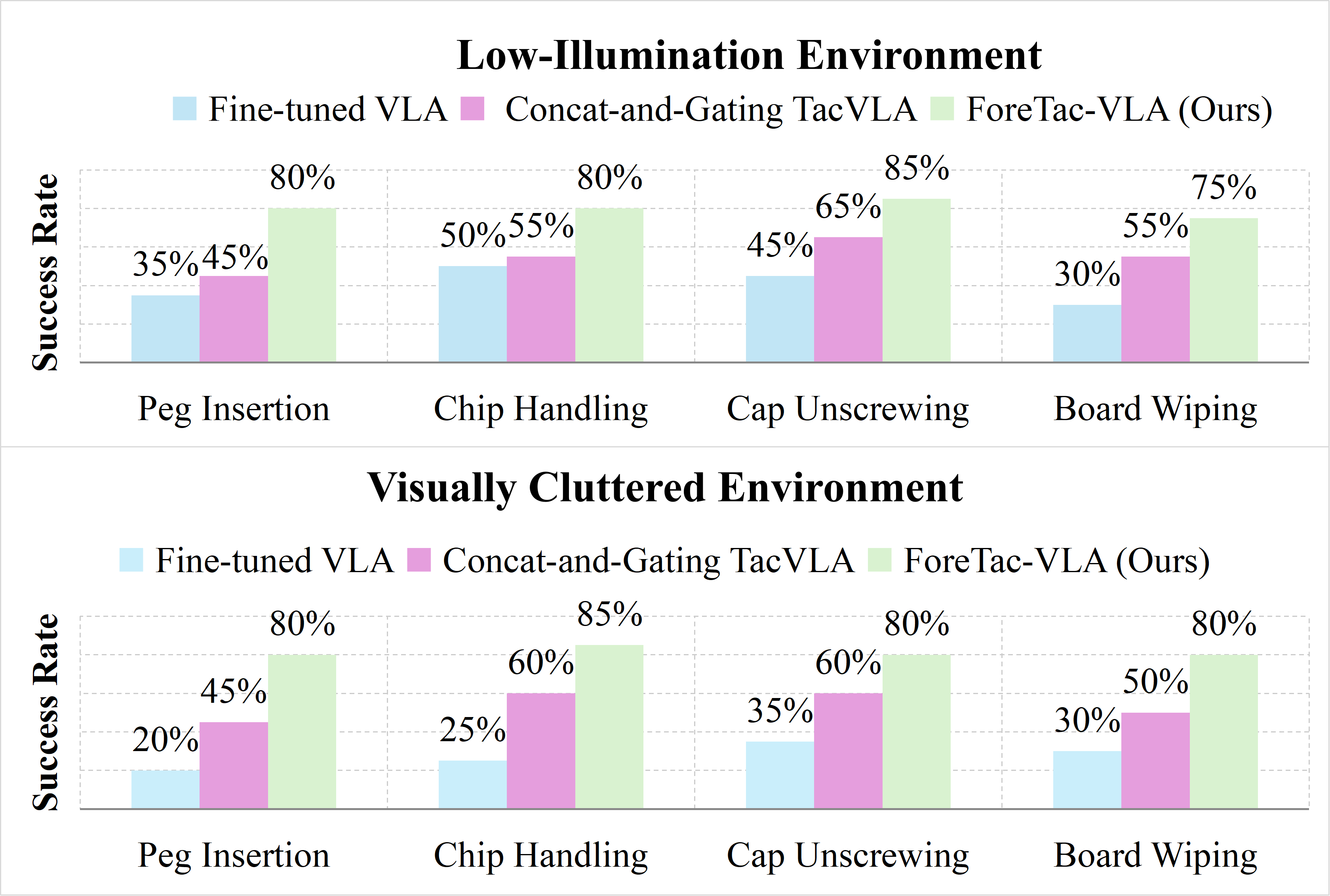}
    \caption{Success rates of the three models under challenging visual conditions. Top: Low-Illumination Environment. Bottom: Visually Cluttered Environment.}
    \label{fig:robustness_results}
\end{figure}

\textbf{2) Effect of Future Tactile Forecasting:} 
Adding future tactile forecasting to the single-direction cross-attention model increases the average success rate from 80.00\% to 81.25\%. 
To directly evaluate the forecasting capability, we compute the mean absolute error (MAE) between the predicted and ground-truth future tactile states on a held-out set. This error is computed in the normalized tactile space and averaged over all future horizons and taxels from both tactile sensors, resulting in an MAE of $6.48 \times 10^{-3}$. 
Fig.~\ref{fig:tactile_forecasting} further shows that the predicted tactile patterns capture the evolving spatial contact distribution over the forecasting horizon. 
These results suggest that future tactile forecasting provides additional anticipatory contact information beyond observed tactile history.

\textbf{3) Effect of Bidirectional Cross-Attention:}
Replacing single-direction cross-attention with bidirectional cross-attention while retaining future tactile forecasting increases the average success rate from 81.25\% to 95.00\%. This substantial improvement demonstrates that two-way interaction between tactile and vision-language representations provides more informative multimodal representations for both future tactile forecasting and action generation.

Overall, tactile input, future tactile forecasting, and bidirectional cross-attention jointly improve multimodal interaction and enable more reliable action generation in contact-rich manipulation.

\subsection{Robustness and Generalization}

To further evaluate the robustness of ForeTac-VLA under challenging visual conditions, we compare ForeTac-VLA with the fine-tuned VLA model and the concat-and-gating TacVLA model in low-illumination and visually cluttered environments. Fig.~\ref{fig:robustness_env} illustrates the two evaluation environments across the four manipulation tasks. Each model is evaluated over 20 trials per task, and Fig.~\ref{fig:robustness_results} reports their success rates under both challenging conditions.

\textbf{1) Low-Illumination Environment:}
In the low-illumination environment, ForeTac-VLA achieves an average success rate of 80.0\%, compared with 55.0\% for the concat-and-gating TacVLA model and 40.0\% for the fine-tuned VLA model. Relative to the normal experimental setting, the average success rate drops for ForTac-VLA, the concat-and-gating TacVLA model, and the fine-tuned VLA model are 15.00, 17.50, and 18.75 percentage points, respectively, indicating that ForeTac-VLA maintains more stable performance under reduced illumination.

\textbf{2) Visually Cluttered Environment:}
In the visually cluttered environment, ForeTac-VLA maintains an average success rate of 81.25\%, compared with 53.75\% for the concat-and-gating TacVLA model and 27.50\% for the fine-tuned VLA model. The average success rate drops for ForTac-VLA, the concat-and-gating TacVLA model, and the fine-tuned VLA model are 13.75, 18.75, and 31.25 percentage points, respectively. For example, on \textit{Peg Insertion}, ForeTac-VLA achieves 80\% success, compared with 45\% and 20\% for the concat-and-gating TacVLA model and the fine-tuned VLA model, respectively.

Overall, ForeTac-VLA achieves the highest success rates and the smallest average performance degradation among the three models in both environments, demonstrating stronger robustness and generalization under visual distribution shifts.

\subsection{Failure Case Analysis}
\begin{figure}[!t]
    \centering
    \includegraphics[width=0.9\columnwidth]{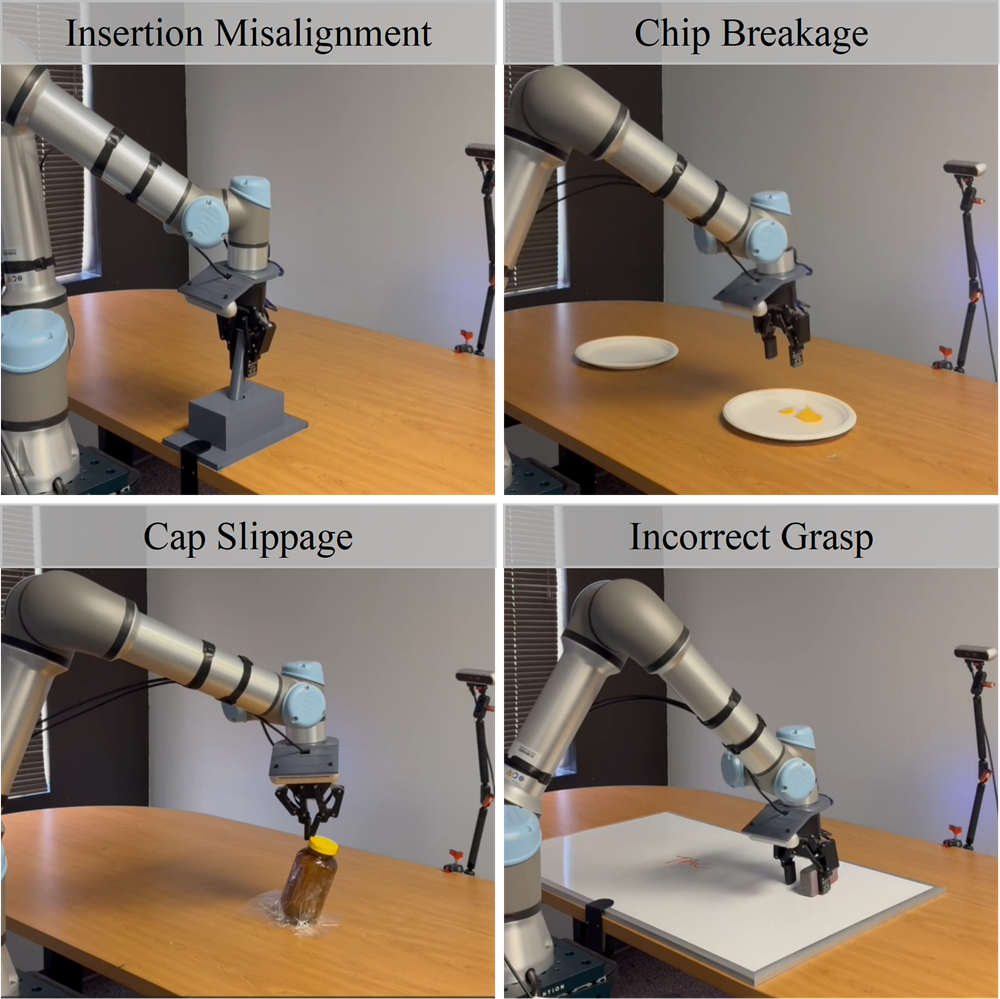}
    \caption{Representative failure cases of the baseline models across four contact-rich manipulation tasks: (a) \textit{Peg Insertion}: insertion misalignment; (b) \textit{Chip Handling}: chip breakage; (c) \textit{Cap Unscrewing}: cap slippage; and (d) \textit{Board Wiping}: incorrect grasp.}
    \label{fig:failure_cases}
    \vspace{-2mm}
\end{figure}

Fig.~\ref{fig:failure_cases} illustrates representative failure modes observed in all the VLA models across the four manipulation tasks. In \textit{Peg Insertion}, insertion misalignment prevents successful placement; in \textit{Chip Handling}, excessive or poorly distributed contact causes chip breakage; in \textit{Cap Unscrewing}, slippage between the gripper and cap interrupts rotational manipulation; and in \textit{Board Wiping}, an incorrect grasp location leads to an unfavorable eraser orientation. These failures highlight the difficulty of maintaining accurate and stable physical interaction in contact-rich manipulation.

\section{LIMITATIONS AND FUTURE WORK}

Despite the strong performance of ForeTac-VLA, several limitations remain. First, our current evaluation is conducted using a single tactile sensing configuration and robot embodiment. Second, the tactile forecaster predicts future tactile states at predefined temporal offsets, which may not fully capture interactions with varying temporal dynamics. Future work will explore tactile representations that generalize across different sensor configurations, adaptive forecasting horizons, and broader evaluations involving longer-horizon tasks and more diverse objects.

\section{CONCLUSION}

We presented ForeTac-VLA, a forecasting-based tactile-vision-language-action model for contact-rich robotic manipulation. By combining bidirectional cross-attention with multi-step future tactile forecasting, ForeTac-VLA integrates observed tactile measurements and predicted future tactile states into action generation. Experiments on four real-world contact-rich manipulation tasks show that ForeTac-VLA achieves an average success rate of 95\%, which consistently outperforms the baseline models and maintains strong performance under degraded visual conditions. These results demonstrate the effectiveness of ForeTac-VLA for reliable contact-rich manipulation under diverse visual conditions.

\bibliographystyle{IEEEtran}
\bibliography{references}

\end{document}